\documentclass{article}
\usepackage{spconf,amsmath,graphicx}
\usepackage{times}
\usepackage{helvet}
\usepackage{xcolor}  
\usepackage{courier}
\usepackage{caption}
\usepackage{booktabs}  
\usepackage{tikz}
\usepackage{multirow}
\usepackage{graphicx}
\usetikzlibrary{positioning}

\usepackage{wrapfig}

\usepackage{algorithmic, algorithm, setspace}

\usepackage{hyperref}
\usepackage{mathtools}
\usepackage{amsmath}
\usepackage{amssymb}
\usepackage{bbm}
\DeclareMathOperator*{\argmax}{arg\,max}

\usepackage{mathtools}
\usepackage{amsmath}
\usepackage{amssymb}
\usepackage{bbm}
\usepackage{color}
\usepackage{subfigure}
\usepackage{caption}
\usepackage{float}
\usepackage{bm}
\usepackage{url}
\usepackage{tikz}
\usepackage{soul}
\usepackage{booktabs}
\DeclareMathSymbol{\shortminus}{\mathbin}{AMSa}{"39}
\usepackage[title]{appendix}
\usepackage{amsfonts}
\usepackage{bbm}
\usepackage[margin=0.7in]{geometry}
\usepackage{soul}
\usepackage{graphicx}

\usepackage{algorithmic, algorithm, setspace}
\newcommand{\parti}{\mathcal{B}}
\definecolor{antiquefuchsia}{rgb}{0.57, 0.36, 0.51}
\definecolor{orange}{rgb}{1.00, 0.647, 0.00}
\definecolor{ao(english)}{rgb}{0.0, 0.5, 0.0}
\definecolor{bleudefrance}{rgb}{0.19, 0.55, 0.91}
\definecolor{britishracinggreen}{rgb}{0.0, 0.26, 0.15}
\newcommand{\setA}{\mathcal{A}}
\title{Evaluation of Categorical Generative Models - Bridging the Gap Between Real and Synthetic Data}
\name{Florence Regol, Anja Kroon, Mark Coates}
\address{Dept. of Electrical and Computer Engineering, McGill University, Montr{\'e}al, Qu{\'e}bec, Canada
}
\begin{document}
%\ninept
%
\maketitle
\begin{abstract}
The machine learning community has mainly relied on  real data to benchmark algorithms as it provides compelling evidence of model applicability.  Evaluation on synthetic datasets can be a powerful tool to provide a better understanding of a model's strengths, weaknesses and overall capabilities. Gaining these insights can be particularly important for generative modeling as the target quantity is completely unknown. Multiple issues related to the evaluation of generative models have been reported in the literature. We argue those problems can be avoided by an evaluation based on ground truth. General criticisms of synthetic experiments are that they are too simplified and not representative of practical scenarios. As such, our experimental setting is tailored to a realistic generative task. We focus on categorical data and introduce an appropriately scalable evaluation method. Our method involves tasking a generative model to learn a distribution in a high-dimensional setting. We then successively bin the large space to obtain smaller probability spaces where meaningful statistical tests can be applied.
We consider increasingly large probability spaces, which correspond to increasingly difficult modeling tasks, and compare the generative models based on the highest task difficulty they can reach before being detected as being too far from the ground truth. We validate our evaluation procedure with synthetic experiments on both synthetic generative models and current state-of-the-art categorical generative models. 
\end{abstract}

\vspace{-1em}
\section{Introduction}
\label{sec:intro}
% 1 - Trash log-likelihood as an evaluation metric
In the machine learning community, evaluation of generative models is an ongoing topic of research~\cite{garbacea2019,celikyilmaz2020,zhou2019,borji2019,thompson2022on,theis2016a,wu2017, nagarajan2021}.
The most common method is to evaluate the (usually upper bounded) log-likelihood on held out test data; the model with the highest log-likelihood is declared the better model. Although principled, this method of evaluation has some known drawbacks as highlighted in~\cite{theis2016a}. A simple example from~\cite{oord2014} shows how good likelihood models can generate poor samples. Simply put, the learned model is far from the true distribution.

Another observation, possibly symptomatic of this issue, is the out-of-distribution (OOD) problem. A notable finding by~\cite{nalisnick2018} is that high likelihood and good sample generation do not guarantee good OOD detection. The task of OOD detection is an important application of generative modeling. If a model is an appropriate approximation of the true distribution, then it should be able to detect low-probability samples.
%Once again, the learned model is likely far from the true distribution.

% 2 - In lit, one approach to address this problem is to report more metrics.
One approach taken in the literature to mitigate this problem is to adopt a more comprehensive and task-oriented methods of evaluation. By assessing over multiple metrics and designing task-oriented metrics, one can better assess the true capabilities of a generative model~\cite{theis2016a,zhou2019,caccia2020}. Nonetheless, this remains a heuristic approach that is unavoidably tied to the application at hand and therefore cannot be applied to the general problem of generative model evaluation.
\begin{figure}[t]
    \centering
    \includegraphics[scale=0.23]{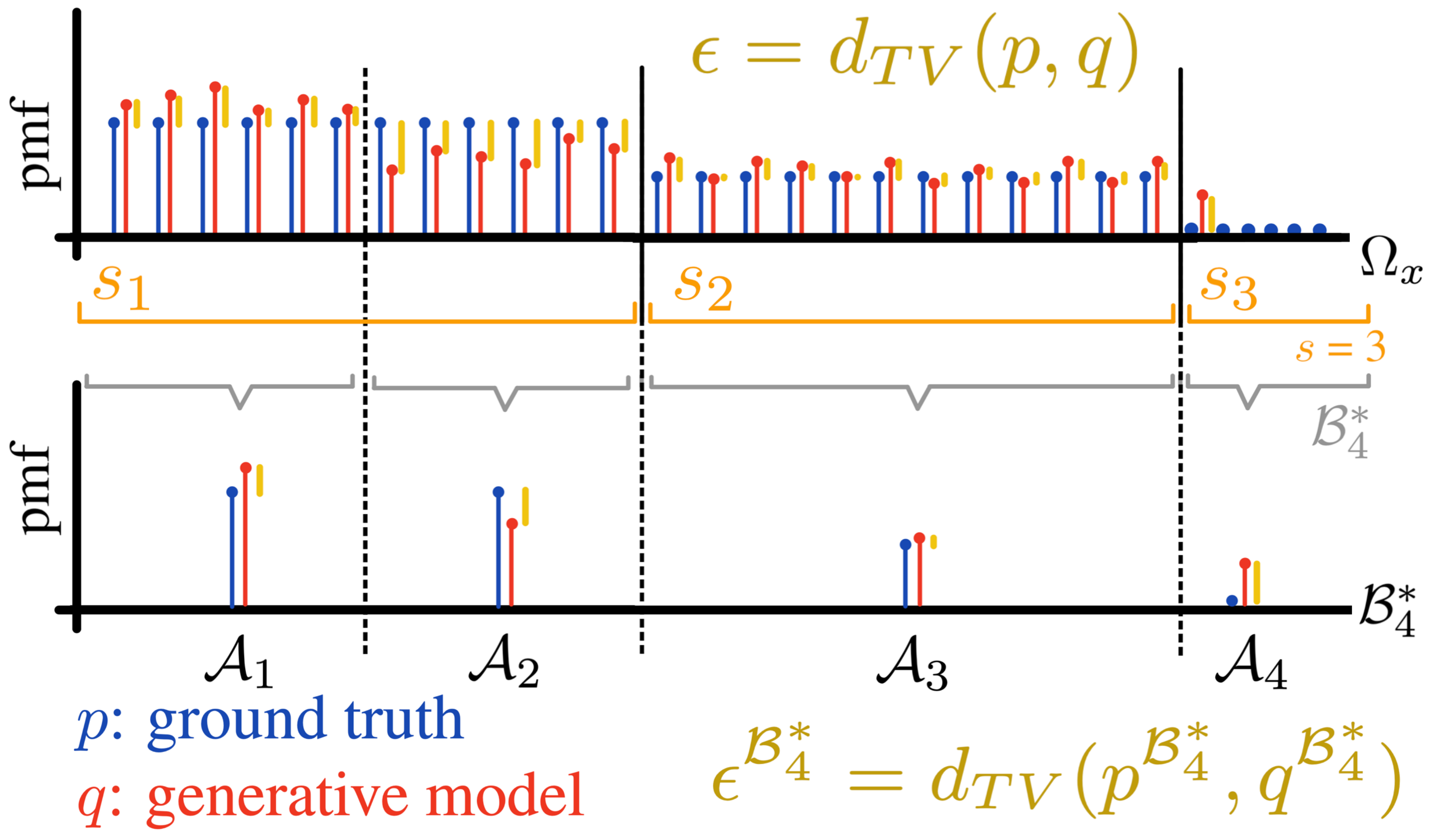}
    \caption{Overview of our proposed binning procedure. The initial distribution $p$ has $s=3$ flat regions on the $\Omega$ space contained in $\mathcal{\mathbf{S}} = \{ \mathcal{S}_1, \mathcal{S}_2, \mathcal{S}_3 \} $ . For a granularity level $k=4$, our binning procedure finds that cutting $\mathcal{S}_1 = \{ \setA_1, \setA_2 \}$ to form $\parti_4^* = \{\setA_1, \setA_2,\setA_3, \setA_4\}$ maximises $d_{TV}(p^{\parti_4},q^{\parti_4}) $. Hypothesis testing can then be conducted over the induced distributions $p^{\parti^*_4},q^{\parti^*_4}$ on the 4 event probability space $\parti_4^*$.}
    \label{fig:overview}
    \vspace{-1em}
\end{figure}

%Using many metrics and designing task-oriented metrics is an way to manage conflicting results from evaluation metrics. 

% 3 - we want to return to synthetic experiment, explain how it alleviates problems briefly
In this work, we propose to return to a synthetic setting where the evaluation of a generative model can be framed as a statistical identity testing problem. This allows us to draw from statistical testing literature. This approach alleviates the problems associated with log-likelihood evaluation. Rather than testing whether high log-likelihood is assigned to unseen samples, we can directly evaluate whether the model assigns the correct probability mass to each element in the space. We can then directly answer questions about OOD capability, sample generation quality, and overall generative modeling performance. This approach also provides better insight into which types of distributions a particular generative model can fit. 
%Many application rely on accurate estimation of prob. space : uncertainty quantification, active learning, calibration, sample generation]
% 4 - Tie our setting and evaluation scheme to realistic scale of experiment

The synthetic task must be representative of the actual task a generative model would be called to solve. In almost all cases, generative models in machine learning target a (very) high dimensional setting. One challenge induced by such settings is that statistical tests based on samples become meaningless. We simply cannot collect enough samples for a test to reject a null hypothesis. In this work, to address this issue, we propose an algorithm to bin the probability space at various granularity levels. We then consider the induced distributions on the smaller probability spaces as a testing proxy for our generative model quality, observing that the error for the induced distributions is a lower bound on the error in the original space. Figure~\ref{fig:overview} provides an overview of our procedure. Random binning is not successful and we explain how we construct bins to maximize the lower bound on the error.

We validate our binning procedure through synthetic experiments, ensuring it can preserve the correct ranking of synthetic generative models. We then show that our testing procedure correctly ranks state-of-the-art generative models trained on our synthetic task by comparing with the empirical total variation on the initial space $\Omega$ and with visual inspection of the learned distributions.

\section{Methodology}
\label{sec:format}
We aim to design an evaluation method or metric that can rank generative models reliably for a \emph{very large scale} discrete distribution space $ |\Omega |$, \emph{given the ground truth distribution} $p$. Given samples from  unknown distributions $q_1$, $q_2$, we wish to distinguish which distribution is closer to a reference distribution $p$ (thus achieving a ranking of generative models). 
\vspace{-1em}
\subsection{Distribution Testing} This setting is analogous to a fundamental problem in statistical distribution testing: identity testing. In this setting, given sample access to an unknown distribution $q$, an identity test declares if $q$ is $\epsilon$-close to a reference distribution $p$ with probability at least $1-\delta$. See~\cite{batu2001,diakonikolas2021} for a more formal definition and~\cite{canonne2020} for a review on this topic. In this field, closeness of discrete distributions $p,q$ on a discrete sample space $\Omega$ is formalized as the total variation distance:
 \begin{align}
   \epsilon= d_{TV}(p,q) & \triangleq \frac{1}{2}||p-q||_1 =  \frac{1}{2} \sum_{x \in \Omega} |p_x -q_x| 
\end{align}
(with $p_x$ used as a shorthand for $p(x)$).
The $\ell^2$ or Hellinger distance $d^2(p,q) \triangleq ||p-q||_2 = \sum_{x \in \Omega} (p_x -q_x)^2  $ can also be used.
For relatively small sized $\Omega$, results from the identity testing literature can be applied directly. However, we target large $\Omega$ where the practical number of samples is  $m \ll |\Omega|$. At this scale, algorithms for identity testing are unusable; the provably most powerful test  has a  $\sqrt{|\Omega|}$ term in its required number of samples~\cite{diakonikolas2021}. 
\vspace{-1em}
\subsection{Binning the space}
We propose to move from the probability space $\Omega$ to a smaller space $\parti$ by binning the elements $x$ in partitioning sets of $\Omega$: $\parti = \{\setA_1, \setA_2,\dots\}$ with $\cup_i \setA_i = \Omega$ and $ \cap \setA_i = \emptyset $. We then assess the distributions induced by this binning to evaluate the models. 

We denote all partitions of $\Omega$ by $\rho(\Omega)$.
The `binned' distribution of the ground truth $p$ is $p^{\parti}$, and the `binned' distribution of the generative model $q$ becomes $q^{\parti}$, where $p^{\parti}_{\setA} = \sum_{x \in \setA} p_x$. This binning operation naturally introduces  a new total variation error on the $\parti$ space 
:
{\small \begin{align}
\epsilon^{\parti} = d_{TV}(p^{\parti},q^{\parti})=&   \frac{1}{2} \sum_{\setA \in \parti } |p^{\parti}_{\setA} - q^{\parti}_{\setA}| =  \frac{1}{2}\sum_{\setA \in \parti } |\sum_{x \in \setA} p_x -q_x | .
\end{align}}
Binning also introduces an error between the initial distribution $p$ and its binned version $p^{\parti}$ that can be defined as:
\begin{align}
    \epsilon^{p \rightarrow p^{\parti}} = \frac{1}{2}\sum_{\setA \in \parti } \sum_{x \in \setA} |p_x - \frac{p^{\parti}_{\setA}}{|\setA|}| .
\end{align}
At the cost of introducing error, binning allows us to reduce the space to a workable size, so that we can use tools from the distribution testing literature. In particular, we can design a hypothesis test that aims to reject $\mathcal{H}_0 : d(p^{\parti},q^{\parti}) < \epsilon$ at a probability significance level $\delta$.

We can see that $d_{TV}(p^{\parti},q^{\parti})$ is a lower bound for $d_{TV}(p,q)$, but the result of such test in the binned space \emph{does not}  provide an direct indication of the outcome in the original space. The intuition is that, if the test fails for some binning $\parti$, this suggests that a generative model is poor, in the sense that it cannot achieve the easier task of representing the binned distribution to within a required error. In the following section, we describe how we select the binning.
\vspace{-1em}
\subsubsection{Choosing the bins}
When moving from $p,q$ to $p^{\parti}, q^{\parti}$, information  is inevitably lost. Additionally, a smaller numbers of bins $k = |\parti|$ will result in greater information loss. Therefore, for a given granularity level $k$,  bins are chosen such that the least possible error on $\epsilon^{\parti_k}$ is introduced for the reference distribution $p$. 

\textbf{Minimizing the binning error of $\epsilon^{p \rightarrow p^{\parti}}$}:
We start by considering bins that minimize the error from going to $p$ to $p^{\parti}$.
Denote by $\parti_k^p(\lambda)$ the set of all $k$ partitions of $\Omega$ s.t. $\epsilon^{p \rightarrow p^{\parti}}$ is at most $\lambda$:
\begin{align}
    \parti_k^p(\lambda) = \{\parti; \epsilon^{p \rightarrow p^{\parti}} \leq \lambda, \parti \in \rho^k(\Omega) \},
\end{align}
where $\rho^k(\Omega)$ denotes the set of all  partitions of $\Omega$ of size $k$.

It is easy to test whether a particular $\parti$ lies in $\parti^p_k(\lambda)$ for any $p$, but $\parti^p_k(\lambda)$ may be very large solution space. For the particular case where no error is tolerated $\parti_k^p(\lambda=0)$, however, the space is much smaller (and may be empty). 

We take advantage of our ability to specify the synthetic distribution $p$, and construct $p$ so that the pmf has $s$ unique values and places all elements with a common value in bins $\mathcal{S}_i$ to form the binning $\mathcal{\mathbf{S}} = \{ \mathcal{S}_1, \dots, \mathcal{S}_s \} $. With a judicious choice of $s$, such a reference distribution is sufficient to provide a stern test of generative models. Given a $p$ with $s = |\mathcal{\mathbf{S}}|$ flat regions, we have that :
{\small \begin{align}
    \parti_k^p(0) =  \begin{cases} \emptyset  &\text{ if } s>k\\
   \mathcal{\mathbf{S}} &\text{ if } s=k \\
   \{ \bigcup_{\mathcal{S}_i \in \mathcal{\mathbf{S}}}  \mathcal{P}_i  ;  \mathcal{P}_i \in \rho^{c_i}(\mathcal{S}_i) ,  \sum^s_{i=1} c_i =k  \}  &\text{ if } s<k 
    \end{cases} \label{eq:min_b}
\end{align}
where $c_i \in \mathbb{N}$ is the cardinally of the partitioning that can be made on $\mathcal{S}_i$. The set $\{ \bigcup_{\mathcal{S}_i \in \mathcal{\mathbf{S}}}  \mathcal{P}_i  ; \mathcal{P}_i \in \rho^{c_j}(\mathcal{S}_i) ,  \sum_j c_j =k  \} $ is all possible union of partitions of each flat region $\mathcal{S}_i$, s.t. we end up with the correct number of bins at the end (enforced by $\sum^s_{i=1} c_i =k$). }

\textbf{Maximizing the binning error of $ p^{\parti}, q^{\parti}$}:
By the triangle inequality, we can see that true error $\epsilon$ will always be greater than $\epsilon^{\parti}$. Hence, given the choice of binning from ($ \parti_k^p(0)$), we can select the binning that \textbf{maximizes}  the error between the induced binned distribution $p^{\parti}$ and the induced binned distribution $q^{\parti}$: 
\vspace{-1em}
\begin{align}
   \parti_k^* =  \argmax_{\parti \in  \parti_k^p(0)} d( p^{\parti}, q^{\parti}).
\end{align}

\begin{algorithm}[t]\caption{Highest granularity level $k$ of $q$}
\begin{algorithmic}
\STATE {\bfseries Input:} $p$ with flat regions $s=|\mathcal{S}|$, sampling access from $q$, error threshold $\epsilon_{test}$, significance level $\delta$.
\STATE Sample the unknown distribution $\{x_i\}^{m}_{i=1} ;  x \sim q $.
\FOR{$k \in [s, \dots, 2s] $}
\STATE Solve for $\parti^*_k$.
\STATE Compute $p^{\parti^*_k}$.
\STATE Construct $\hat{q}_{emp}^{\parti^*_k}$ from the $m$ samples.
\STATE Test for $d_{TV}(p^{\parti^*_k}, \hat{q}_{emp}^{\parti^*_k})< \epsilon$ at significance $ \delta $.
\IF {the test FAILS}
\STATE Return the granularity $k$ at which the test failed.
\ENDIF
\ENDFOR
\end{algorithmic} \label{algo:testing_procedure}
\end{algorithm}
This concludes how we select bins at a specific granularity level $\parti_k^*$.
For a given flat region $\setA_i$, if we seek a $k=2$ bin-granularity: $\rho^2(\mathcal{S}_i)$, the unique solution that maximises the error is to split the positive error region $\{ x \in \mathcal{S}_i ; q_x -p_x >0  \}$ from the negative error region $\{ x \in \mathcal{S}_i ; q_x -p_x <0  \}$. If there is no positive or negative error, then every bin is equivalent.
Thus, different and tractable solutions for the granularity $k=\{s, \dots, 2s \}$ can be obtained: 
\vspace{-1em}
\begin{align}
   \parti_k^* =  \argmax_{\parti \in \parti_k^p(0)} d( p^{\parti}, q^{\parti}) \label{eq:max_part}.
\end{align}
We have presented the discussion in terms of $q$ and $q^{\parti}$, but we assume that we only have sample access to $q$ (and hence to $q^{\parti}$). Hence, only the empirical pmf of $q$ is accessible. We can approximate with: $\parti_k^* \approx  \argmax_{\parti \in \parti_k^p(0)} d( p^{\parti}, \hat{q}_{emp }^{\parti})$.

A limitation of this procedure is that it is tied to the number of flat regions $s$ of the distribution $p$. However, since we have control over $p$, we can choose $p$ with a number of flat regions that achieves a trade off between (i) posing a suitably demanding challenge and (ii) generating a test that does not require too many samples. In general, in very high dimensions, such as those of protein sequences, for example, structured distributions can be very challenging to learn even if there are only three unique values. 

\vspace{-1em}
\subsection{Hypothesis testing}
We now have a sequence of binnings of increasingly difficult granularity levels, ranging from $k=s$ to $k=2s$ : $\parti_s^*, \parti_{s+1}^*, \dots, \parti_{2s}^*$. At each granularity level $k$, we can use a hypothesis test to detect poor approximation of the ground truth. Since the sampling complexity of the test scales with the probability space and we control the granularity level $k$, we can easily generate $O(k)$ samples. Our null hypothesis is  that the binned generative model $q^{\parti^*_k}$ is within some error $\epsilon_{test}$ of the ground truth distribution $p^{\parti^*_k}$; i.e. $\mathcal{H}_0: d^2(p^{\parti^*_k}, q^{\parti^*_k})< \epsilon_{test} $. We use the $\ell^2$ distance as it is more practical to test for. If the test is able to reject at some granularity $k$, we stop and return this $k$ as the highest granularity level reached by the generative model $q$. For the hypothesis testing, we use the closeness test from~\cite{batu2013}, which tests the distance between two distributions, both of which only have sample access. To make the test applicable to our context, we simply replace the statistical estimates for unknown $p$ by our known statistics for $p$.  The complete testing procedure is outlined in Algorithm~\ref{algo:testing_procedure}.

\section{Experiments}
\vspace{-1em}
\label{sec:pagestyle}
\subsection{Datasets}

 Our synthetic experiments are designed to emulate real world datasets.
The most important characteristic is that the sampling space is several orders larger than the number of generated samples $m$. Second, the  support of the targeted distribution (that we denote $\Omega^+ = \{ x \in \Omega ; p_x >0\}$) is believed to be much smaller than the whole space, i.e. $\Omega^+ /\Omega \approx 0$. The intuition behind this feature is that uniform random sampling is very unlikely to give a valid sample. This is a characteristic of most applications of high-dimensional generative models. Nonetheless, the positive space is still generally believed to be very large and scales with the size of the space. We choose the ratio of approximately $|\Omega^+|/|\Omega| \approx c!/c^c$ as it loosely approaches an estimate given in~\cite{trinquier2021} of the size of a family’s sequence space.
Since we want to keep the number of flat regions small (as explained in the previous section), we choose to model $p$ by a stair distribution with $s$ stairs, where the last stair is a assigned zero probability mass. Hence for a given $s$, we have
\begin{align}
    p(x) = \begin{cases} 
    p_{\setA_i} \quad& \text{if } x \in \setA_i \quad (i \in \{1,\dots, s-1\} )\\
    0 \quad& \text{if } x \in \setA_s
    \end{cases}.
\end{align}
% For training the categorical generative models, we construct $p$ with $s =3$ by splitting all permutations of the sequence into two sets $\setA_1, \setA_2$ with $p_{\setA_1} = 3*p_{\setA_2}$, and setting the probability mass to $0$ for the non-permutation sequences.

\subsection{Baselines}

\textbf{Synthetic $q$}: Even though we could scale to high $\Omega$, for visualisation purposes, we set $|\Omega| = 6^6$ and generate a smaller number of samples $m=1000$. We define a stair distribution  $p$ with $s=4$ flat regions.  We produce 4 synthetic generative models $q_1,q_2,q_3,q_4$ by increasingly perturbing the ground truth distribution:  $d_{TV}(p,q_1) = 0$, $d_{TV}(p,q_2) = 0.1$,  $d_{TV}(p,q_3) = 0.15$, $d_{TV}(p,q_4) = 0.2 $. 

\textbf{Generative models $q$}: We evaluate our evaluation method on state-of-the-art generative models for categorical data. We set $|\Omega| = 6^6$ and define a stair distribution  $p$ with $s=3$ flat regions.  \textbf{CNF}~\cite{lippe2021categorical} is a normalizing flow method that learns a mapping from the categorical space to a continuous representation. \textbf{CDM}~\cite{hoogeboom2021} is a diffusion-based model that operates in the discrete space. \textbf{argmaxAR}~\cite{hoogeboom2021argmax} is a normalizing flow method that uses an argmax operation to map a continuous representation to the discrete space.  For each trial, we generate $m=10,000$ samples.
\vspace{-1em}

\begin{figure}[t]
    \centering
    \includegraphics[scale=0.55]{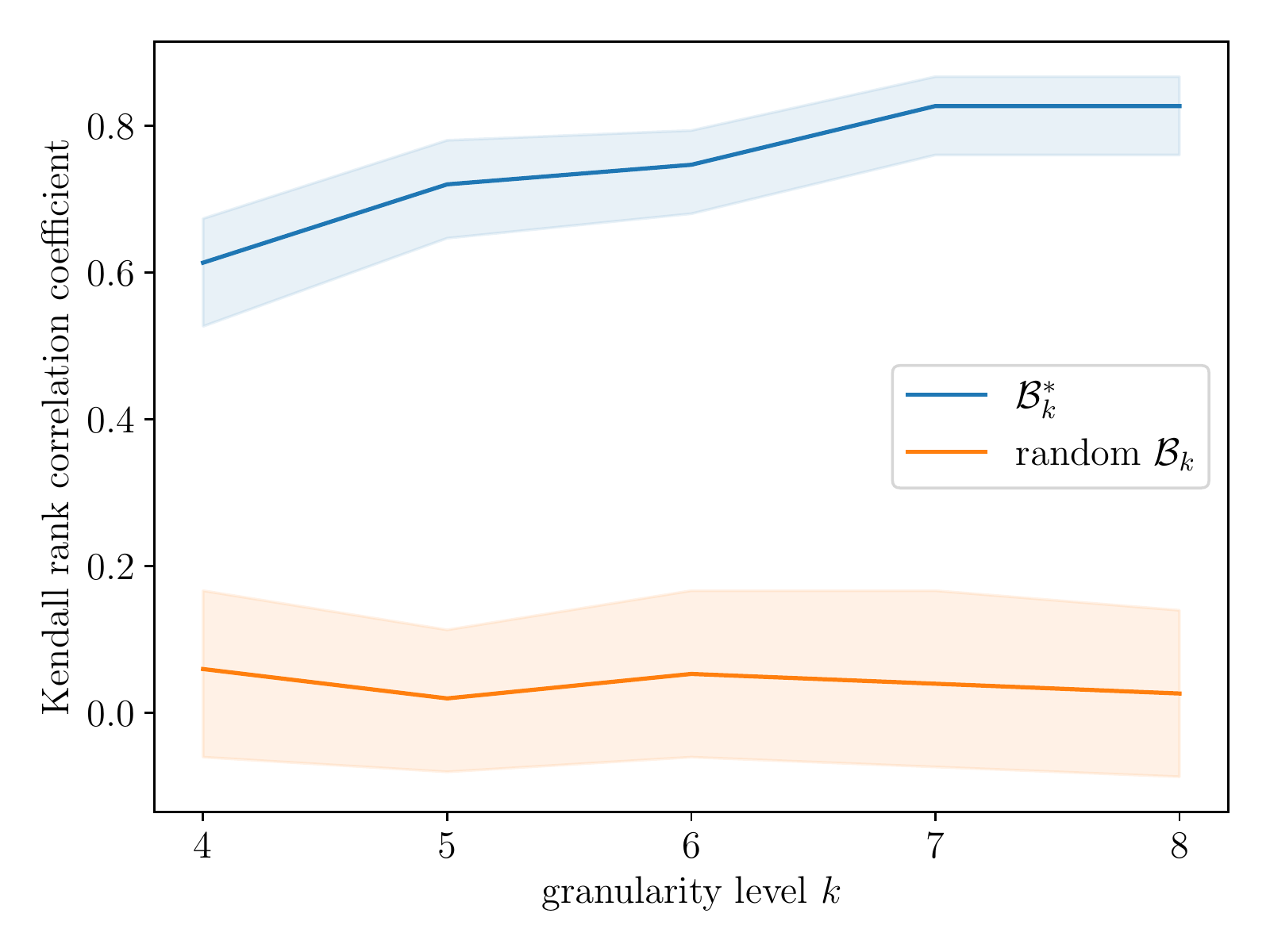}
    \caption{Accurate ranking measure (Kendall Tau) of our proposed binning algorithm vs a random binning baseline. The proposed algorithm is able to correctly rank the  baselines consistently and outperforms a random binning. We report the mean over $50$ trials with a $90\%$ bootstrapped CI.   }
    \label{fig:ranking}
     \vspace{-1em}
\end{figure}
\begin{figure}[H]
  \centering
  \centerline{\includegraphics[width=8.5cm]{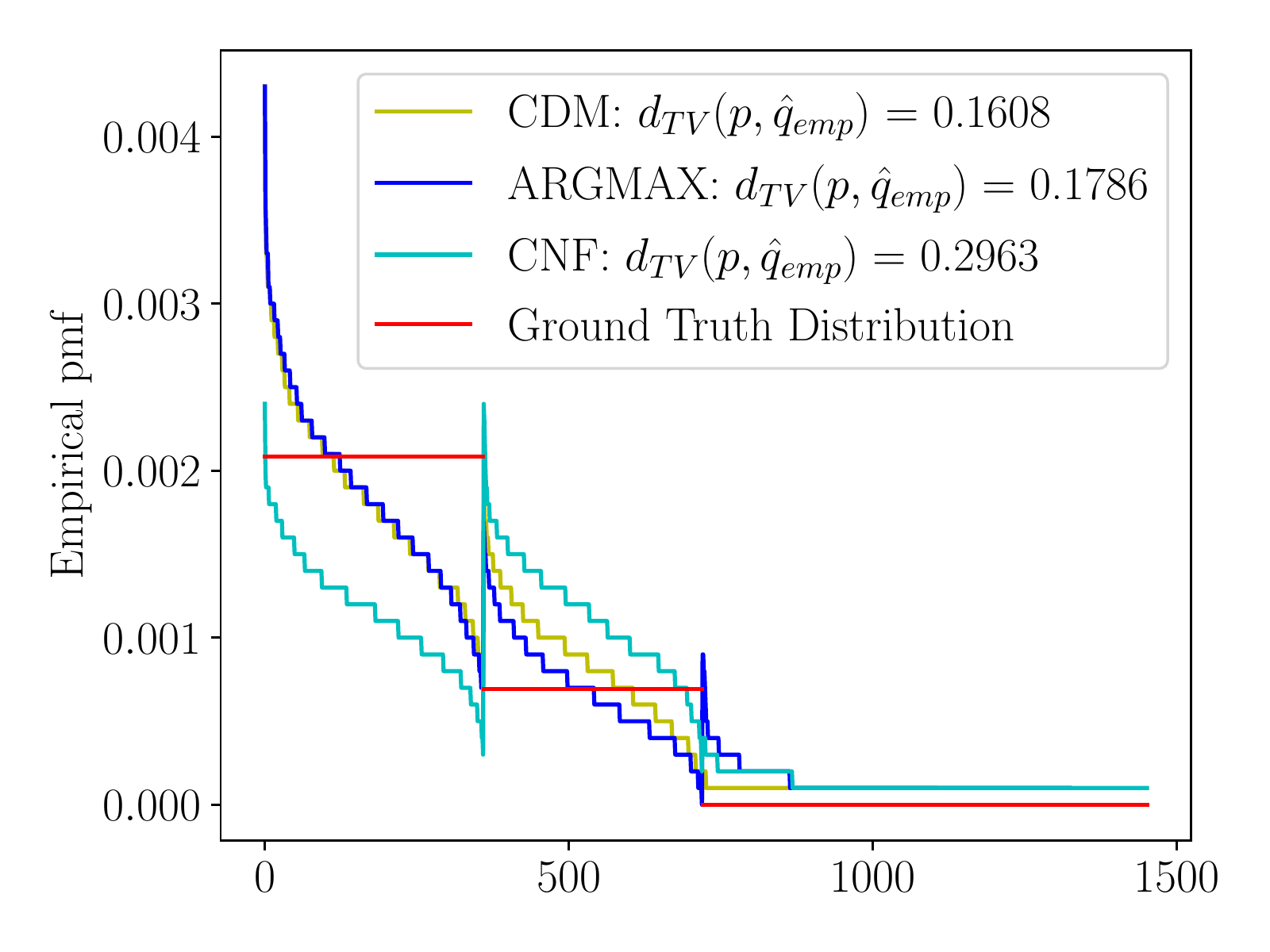}}
  \caption{Empirical pmf of the CDM, argmaxAR, and CNF models over a subset of $\Omega$. The ground truth $p$ is shown in red. The x-axis is sorted differently for each models. For a given model, within a flat region, we sort the $x \in \setA_i$ on the x-axis from the most overestimating $\hat{q}^{emp}_x$ to the most underestimating $\hat{q}^{emp}_x$ w.r.t. ground truth. For each model, we compute the empirical total variation error $d_{TV}(p,\hat{q}^{emp})$. }
  \label{fig:emp}
   \vspace{-1em}
\end{figure}
\subsection{Experiment Details} \label{sec:exp_details}
We begin by validating our algorithm for choosing $\parti^*_k$. We verify the ordering of the empirical total variation given by $\epsilon^{\parti}_{emp} \triangleq d_{TV}(p^{\parti},\hat{q}^{\parti}_{emp}) $ 
of the synthetically created $q_1,q_2,q_3,q_4$ aligns with the correct ranking: $q_1< q_2 < q_3 < q_4$. We compare the ordering provided by the empirical total variation derived form our binning algorithm  $\parti^*_k$ $d_{TV}(p^{\parti^*_k},\hat{q}^{\parti^*_k}_{emp})$
with the empirical total variation derived from some random binning $\parti_k$  $d_{TV}(p^{\parti_k},\hat{q}^{\parti_k}_{emp})$. To compare raking accuracy, we compute the Kendall Tau correlation coefficient which measures the piece-wise displacements between two rankings.

A higher value corresponds to a more closely matched rank. Results can be viewed in Figure~\ref{fig:ranking}.
%\vspace{-1em}
\begin{figure}[t] %[bth]
    \centering
    \includegraphics[scale=0.3]{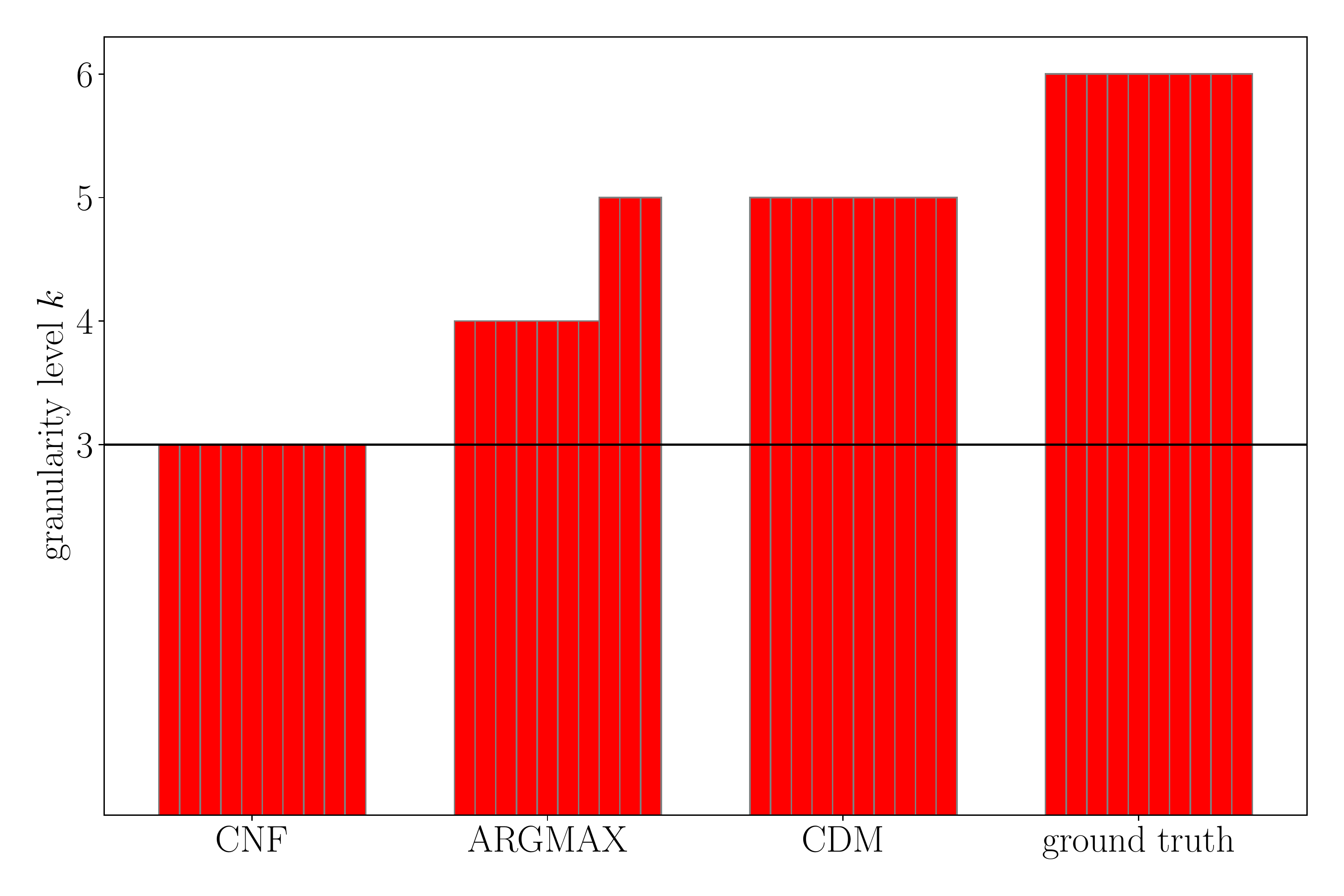}
    \caption{Highest granularity level $k$ obtained by the generative models for 10 trials.  }
    \label{fig:hypothesis}
    % \vspace{-1em}
\end{figure}
Next, we test our evaluation procedure outlined in Algorithm~\ref{algo:testing_procedure} on state-of-the-art categorical generative models. We validate that the generative model reaching the highest granularity level $k$ with our test is the generative model with the lowest empirical total variation on $\Omega$ $d_{TV}(p,\hat{q}^{emp})$.  We set the significance level of the hypothesis testing to $\delta = 0.05$ and test for $\epsilon_{test} = 0.1$. We train each model on a dataset of $10,000$ training samples generated from $p$ for 1000 epochs. For each baseline, a hyperparameter search is conducted over the hidden size = $[8,\dots 64]$, depth = $[1,\dots, 4]$ and learning rate $=[1e-1, \dots, 1e-5]$ and we set the remaining parameters to the default values provided in each respective paper. 
\vspace{-1em}

\section{Discussion}

Figure~\ref{fig:ranking} shows the binning procedure we propose is sensible. In Figure~\ref{fig:emp}, we show the empirical distribution generated from one trial of the generative models that were trained on the synthetic dataset as described in Section~\ref{sec:exp_details}. Visually, we see that the CNF baseline distribution is a very poor approximation of the ground truth distribution. This is also reflected in its empirical total variation error $d_{TV}(p,\hat{q}^{emp})$. The argmaxAR and CDM baselines are closer, with the CDM slightly outperforming argmaxAR based on $d_{TV}(p,\hat{q}^{emp})$. This is reflected in our hypothesis testing ranking in Figure~\ref{fig:hypothesis}, where we see that all trials of the CDM pass the test at $k=4$, whereas only a fraction of the argmaxAR trials pass the same test. Every CNF trial is rejected.
\vspace{-1em}

\section{Conclusion} 
In conclusion, we have introduced an alternative way to evaluate generative models for categorical data. Even though our approach is based on statistical tests, it remains applicable for very large distribution space which is the setting of interest for those models. In future work, we plan on providing theoretical guarantee of our procedure and to further exploit our control over $p$ to design an even more powerful test. 
pagebreak

\section{COPYRIGHT FORMS}
© 2023 IEEE. Personal use of this material is permitted. Permission from IEEE must be obtained for all other uses, in any current or future media, including reprinting/republishing this material for advertising or promotional purposes, creating new collective works, for resale or redistribution to servers or lists, or reuse of any copyrighted component of this work in other works.
% Below is an example of how to insert images. Delete the ``\vspace'' line,
% uncomment the preceding line ``\centerline...'' and replace ``imageX.ps''
% with a suitable PostScript file name.
% -------------------------------------------------------------------------
%\begin{figure}[htb]

%\begin{minipage}[b]{1.0\linewidth}
%  \centering
%  \centerline{\includegraphics[width=8.5cm]{image1}}
%  \vspace{2.0cm}
%  \centerline{(a) Result 1}\medskip
%\end{minipage}
%
%\begin{minipage}[b]{.48\linewidth}
%  \centering
%  \centerline{\includegraphics[width=4.0cm]{image3}}
%  \vspace{1.5cm}
%  \centerline{(b) Results 3}\medskip
%\end{minipage}
%\hfill
%\begin{minipage}[b]{0.48\linewidth}
%  \centering
%  \centerline{\includegraphics[width=4.0cm]{image4}}
%  \vspace{1.5cm}
%  \centerline{(c) Result 4}\medskip
%\end{minipage}
%
%\caption{Example of placing a figure with experimental results.}
%\label{fig:res}
%
%\end{figure}

% To start a new column (but not a new page) and help balance the last-page
% column length use \vfill\pagebreak.
% -------------------------------------------------------------------------
%\vfill
%\

\bibliographystyle{IEEE}
\bibliography{ref}

\begin{thebibliography}{10}

\bibitem{garbacea2019}
Cristina Garbacea, Samuel Carton, Shiyan Yan, and Qiaozhu Mei,
\newblock ``Judge the judges: A large-scale evaluation study of neural language
  models for online review generation,''
\newblock in {\em Proc. Conf. on Empirical Methods in Natural Language Process.
  and Int. Joint Conf. on Natural Language Process EMNLP-IJCNLP}, 2019.

\bibitem{celikyilmaz2020}
Asli Celikyilmaz, Elizabeth Clark, and Jianfeng Gao,
\newblock ``Evaluation of text generation: A survey,''
\newblock arXiv preprint: arXiv 2006.14799, 2020.

\bibitem{zhou2019}
Sharon Zhou, Mitchell~L. Gordon, Ranjay Krishna, Austin Narcomey, Li~Fei{-}Fei,
  and Michael~S. Bernstein,
\newblock ``{HYPE:} {A} benchmark for human eye perceptual evaluation of
  generative models,''
\newblock in {\em Proc. Adv. Neural Info. Process. Syst. NeurIPS}, 2019.

\bibitem{borji2019}
Ali Borji,
\newblock ``Pros and cons of gan evaluation measures,''
\newblock {\em Computer Vision and Image Understanding}, vol. 179, pp. 41--65,
  2019.

\bibitem{thompson2022on}
Rylee Thompson, Boris Knyazev, Elahe Ghalebi, Jungtaek Kim, and Graham~W.
  Taylor,
\newblock ``On evaluation metrics for graph generative models,''
\newblock in {\em Proc. Int. Conf. Learning Representations ICLR}, 2022.

\bibitem{theis2016a}
L.~Theis, A.~van~den Oord, and M.~Bethge,
\newblock ``A note on the evaluation of generative models,''
\newblock in {\em Proc. Int. Conf. Learning Representations ICLR}, 2016.

\bibitem{wu2017}
Yuhuai Wu, Yuri Burda, Ruslan Salakhutdinov, and Roger~B. Grosse,
\newblock ``On the quantitative analysis of decoder-based generative models,''
\newblock in {\em Proc. Int. Conf. Learning Representations ICLR}, 2017.

\bibitem{nagarajan2021}
Vaishnavh Nagarajan, Anders Andreassen, and Behnam Neyshabur,
\newblock ``Understanding the failure modes of out-of-distribution
  generalization,''
\newblock in {\em Proc. Int. Conf. Learning Representations ICLR}, 2021.

\bibitem{oord2014}
Aaron van~den Oord and Benjamin Schrauwen,
\newblock ``Factoring variations in natural images with deep gaussian mixture
  models,''
\newblock in {\em Proc. Adv. Neural Info. Process. Syst. NeurIPS}, 2014.

\bibitem{nalisnick2018}
Eric Nalisnick, Akihiro Matsukawa, Yee~Whye Teh, Dilan Gorur, and Balaji
  Lakshminarayanan,
\newblock ``Do deep generative models know what they don't know?,''
\newblock in {\em Proc. Int. Conf. Learning Representations ICLR}, 2019.

\bibitem{caccia2020}
Massimo Caccia, Lucas Caccia, William Fedus, Hugo Larochelle, Joelle Pineau,
  and Laurent Charlin,
\newblock ``Language gans falling short,''
\newblock in {\em Proc. Int. Conf. Learning Representations ICLR}, 2020.

\bibitem{batu2001}
T.~Batu, E.~Fischer, L.~Fortnow, R.~Kumar, R.~Rubinfeld, and P.~White,
\newblock ``Testing random variables for independence and identity,''
\newblock in {\em Proc. IEEE Symp. on Foundations of Computer Sci.}, 2001.

\bibitem{diakonikolas2021}
Ilias Diakonikolas, Themis Gouleakis, Daniel~M. Kane, John Peebles, and Eric
  Price,
\newblock ``Optimal testing of discrete distributions with high probability,''
\newblock in {\em Proc. ACM SIGACT Symp. on Theory of Comput.}, 2021.

\bibitem{canonne2020}
Cl{\'{e}}ment~L. Canonne,
\newblock {\em A Survey on Distribution Testing: Your Data is Big. But is it
  Blue?},
\newblock Number~9 in Graduate Surveys. 2020.

\bibitem{batu2013}
Tu\u{g}kan Batu, Lance Fortnow, Ronitt Rubinfeld, Warren~D. Smith, and Patrick
  White,
\newblock ``Testing closeness of discrete distributions,''
\newblock {\em J. ACM}, vol. 60, no. 1, 2013.

\bibitem{trinquier2021}
Jeanne Trinquier, Guido Uguzzoni, Andrea Pagnani, Francesco Zamponi, and Martin
  Weigt,
\newblock ``Efficient generative modeling of protein sequences using simple
  autoregressive models,''
\newblock {\em Nature Communications}, vol. 12, 2021.

\bibitem{lippe2021categorical}
Phillip Lippe and Efstratios Gavves,
\newblock ``Categorical normalizing flows via continuous transformations,''
\newblock in {\em Proc. Int. Conf. Learning Representations ICLR}, 2021.

\bibitem{hoogeboom2021}
Emiel Hoogeboom, Didrik Nielsen, Priyank Jaini, Patrick Forr\'{e}, and Max
  Welling,
\newblock ``Argmax flows and multinomial diffusion: Learning categorical
  distributions,''
\newblock in {\em Proc. Adv. Neural Info. Process. Syst. NeurIPS}, 2021.

\bibitem{hoogeboom2021argmax}
Emiel Hoogeboom, Didrik Nielsen, Priyank Jaini, Patrick Forr{\'e}, and Max
  Welling,
\newblock ``Argmax flows: Learning categorical distributions with normalizing
  flows,''
\newblock in {\em Proc. Symposium on Adv. in Appr. Bayesian Inference}, 2021.

\end{thebibliography}

%\bibliography{ref} what was in aaai

% References should be produced using the bibtex program from suitable
% BiBTeX files (here: strings, refs, manuals). The IEEEbib.bst bibliography
% style file from IEEE produces unsorted bibliography list.
% -------------------------------------------------------------------------

\end{document}